# Initialization of a Polyharmonic Cascade, Launch and Testing

Bakhvalov Y. N., Ph.D., Independent Researcher, Cherepovets

bahvalovj@gmail.com, ORCID: 0009-0002-5039-2367

**Abstract.**

This paper concludes a series of studies on the polyharmonic cascade – a deep machine learning architecture theoretically derived from indifference principles and the theory of random functions. A universal initialization procedure is proposed, based on symmetric constellations in the form of hyperoctahedra with a central point. This initialization not only ensures stable training of cascades with tens and hundreds of layers (up to 500 layers without skip connections), but also radically simplifies the computations. Scalability and robustness are demonstrated on MNIST (98.3% without convolutions or augmentations), HIGGS (AUC ≈ 0.885 on 11M examples), and Epsilon (AUC ≈ 0.963 with 2000 features). All linear algebra is reduced to 2D operations and is efficiently executed on GPUs. A public repository and an archived snapshot are provided for full reproducibility.

This paper is devoted to the problem of initializing the polyharmonic cascade, a machine learning architecture whose theoretical foundations were laid out in [3], [4], and [5]. The paper also presents the results of tests performed with this architecture.

In [3] it was shown how a machine learning regression problem can be solved within the framework of the theory of random functions [18]. The resulting solution is a particular form of polyharmonic spline [7], [13]. A theoretical connection was established between the polyharmonic spline and random function theory in the context of solving a machine learning regression problem. A mathematical construction was obtained that provides a universal solution to multivariate approximation tasks.

Paper [4] is devoted to scaling the solution obtained in [3]. It was shown that the same mathematical construction is well suited for representing and computing a large number of functions simultaneously. In this setting, however, these functions must be specified by their values at a common set of key points. This specially defined set of points was called a "constellation," and the family of functions represented in this way was called a "package" of polyharmonic splines. It was further justified to connect such packages into a cascade, forming a more complex computational system. Efficient matrix-based procedures were proposed for both forward computation and differentiation.

In [5] a training method for the polyharmonic cascade was proposed, derived as the solution of an optimization problem. The method is not gradient descent, although it does require a differentiation step via the chain rule, moving backward through the packages in the cascade. As a result, [5] describes the polyharmonic cascade and its operating principles – as a machine learning architecture – in the form of a complete, self-contained algorithm.

However, the procedure for initializing the polyharmonic cascade was left outside the scope of the previous works. From a practical point of view this question is extremely important. Moreover, the choice of constellations also matters theoretically: an appropriate choice can simplify the operations being executed and significantly change the computational load, as will be shown below.

To discuss the initialization procedure for the polyharmonic cascade, we first briefly describe how the cascade itself operates and how it is trained. The derivation and justification of all these steps were given in [4] and [5]; here we simply present them in their final form, without repeating the proofs.

We now reproduce the schematic diagram of the cascade from [5].

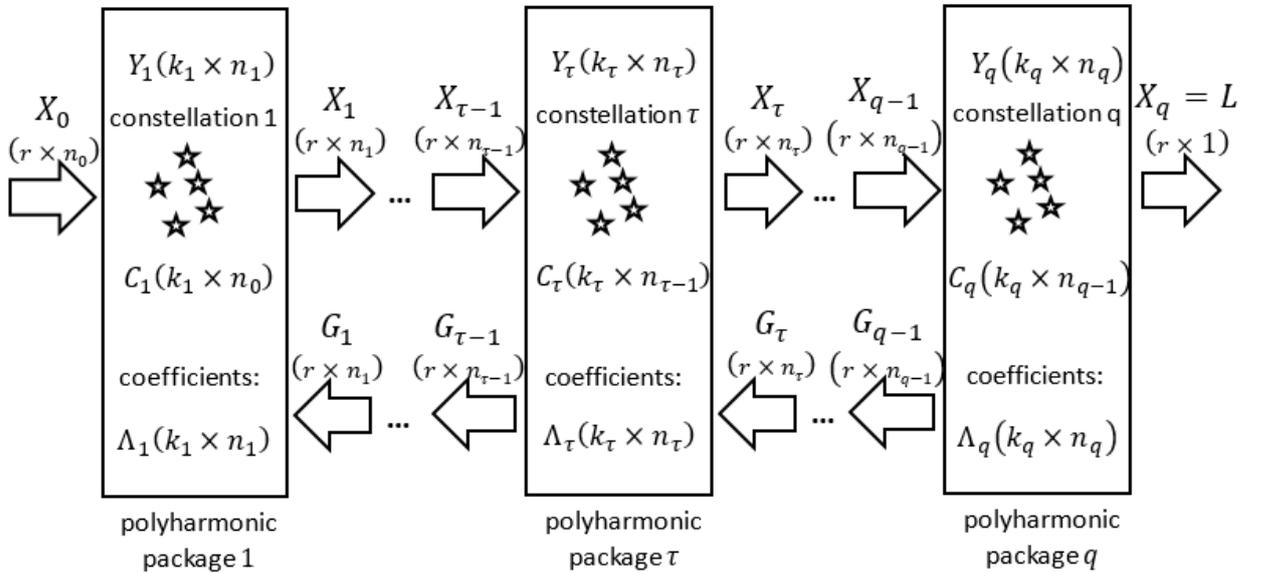

Figure 1. Polyharmonic cascade

Figure 1 shows a polyharmonic cascade consisting of a sequence of packages ($q$ in total). The diagram explicitly depicts the first and last packages, as well as some intermediate package with index $\tau$. The numbers of inputs and outputs of the packages are denoted by $n_0, n_1, \ldots, n_{q-1}$, while the numbers of points in the constellations are denoted by $k_1, k_2, \ldots, k_q$. The number of vectors processed in one batch is denoted by $r$ (the batch size).

Data are processed by the cascade as follows.

At the input of the first package, we receive a batch of data in the form of a matrix $X_0$ of size $(r \times n_0)$. After processing $X_0$, the first package produces an output matrix $X_1$ of size $(r \times n_1)$, which is then fed into the second package, where $X_2$ is obtained, and so on. At the output of the cascade we obtain a matrix $X_q$ of size (r×1), which in [4] and [5] is also denoted by the vector $L$.

From the shape of $L$ $(r \times 1)$, we see that we are considering the case where the cascade has only a single output (the last polyharmonic package computes just one function). In [5] this single-output case was analyzed in detail first, and then variants were given showing how to transform it into a cascade with multiple outputs. For the purposes of the present paper, focusing on cascade initialization, it is sufficient to treat the single-output case (i.e., a cascade computing one function). Extension to multiple outputs does not change anything essential in this context and can be carried out using the same techniques as described in [5].

Each polyharmonic package in the cascade is described by three matrices. For the package with index $\tau$ these are $C_\tau, Y_\tau$, and $\Lambda_\tau$. The matrix $C_\tau$ specifies the constellation points; the matrix $Y_\tau$ specifies the values of the polyharmonic functions (belonging to the package) at those points; and the matrix $\Lambda_\tau$ contains the coefficients of the equations directly used in computation (the matrix $\Lambda_\tau$ can be recovered from the known $C_\tau$ and $Y_\tau$).

To perform a training step, after processing a batch we must move backward through the cascade and sequentially compute the derivative matrices from $G_{q-1}$ down to $G_1$. Then, for each package independently, we compute its corresponding matrix $\Omega$. All matrices $\Omega_1, \Omega_2, \ldots, \Omega_q$ are summed (they all have the same shape $(r \times r)$ for every package). A linear system is then solved to obtain a vector $B$ of length $r$, which is subsequently applied independently to each package in order to update the matrix of function values $Y_\tau$; from the updated $Y_\tau$ we then recompute the coefficient matrix $\Lambda_\tau$.

We now describe all of the above steps in more detail, in the form of matrix operations as derived in [4] and [5].

First, we show how an arbitrary polyharmonic package with index $\tau$ computes the matrix $X_\tau$ from the input matrix $X_{\tau-1}$.

As a first step, we compute the matrix of squared distances between all row vectors of $X_{\tau-1}$ and all points of the constellation. Denote this matrix by $M_\tau$:

$$M_\tau = N_{\tau x} J_{1,k_\tau} + J_{r,1} N_{\tau c}^T - 2 X_{\tau-1} C_\tau^T, \tag{1}$$

where

$$N_{\tau x} = (X_{\tau-1} \circ X_{\tau-1})J_{n_{\tau-1},1},$$

$$N_{\tau c} = (C_\tau \circ C_\tau)J_{n_{\tau-1},1},$$

$\circ$ denotes the Hadamard (elementwise) product,
$J_{1,k_\tau}$ is a $1 \times k_\tau$ row vector of ones,
$J_{r,1}$ is an $r \times 1$ column vector of ones,
$J_{n_{\tau-1},1}$ is an $n_{\tau-1} \times 1$ column vector of ones,
and $C_\tau$ is the constellation matrix of the $\tau$-th package.

Next, from the matrix $M_\tau$ we obtain a matrix $K_\tau$ of the same size $(r \times k_\tau)$ by applying a scalar transformation to each element:

$$k_{ip}^\tau = \frac{1}{2} m_{ip}^\tau \left(\ln(m_{ip}^\tau) - 2b\right) + c, \qquad (2)$$

where
$b$ and $c$ are coefficients whose values were estimated in [3],
$k_{ip}^\tau$ is an element of the matrix $K_\tau$ (not to be confused with $k_\tau$),
$m_{ip}^\tau$ is the corresponding element of $M_\tau$,
$i = \overline{1,r}, p = \overline{1,k_\tau}$.

We then compute $X_\tau$ as the matrix product of $K_\tau$ and $\Lambda_\tau$:

$$X_\tau = K_\tau \Lambda_\tau \qquad (3)$$

Thus, processing data in any package amounts to performing operations (1), (2), and (3). Note that the matrix $Y_\tau$ (which contains the function values at the constellation points) does not appear explicitly here. However, it is precisely the entries of $Y_\tau$ that are the subject of optimization in the training algorithm described in [5].

Knowing $Y_\tau$, we can easily compute $\Lambda_\tau$:

$$\Lambda_\tau = U_\tau Y_\tau \qquad (4)$$

where $U_\tau$ is a square matrix completely determined by the constellation of the package; its size $(k_\tau \times k_\tau)$ is set by the number of constellation points. If the constellation is fixed and does not change during training – as assumed in the training algorithm of [5] – then $U_\tau$ only needs to be computed once, at cascade initialization.

We now describe how to compute $U_\tau$ from a given constellation matrix $C_\tau$.

First, we compute the matrix of all pairwise squared distances between the constellation points themselves; denote it by $M_\tau^{(C)}$:

$$M_\tau^{(C)} = N_{\tau c}J_{1,k_\tau} + J_{k_\tau,1}N_{\tau c}^T - 2C_\tau C_\tau^T ,\qquad(5)$$

where

$$N_{\tau c} = (C_\tau \circ C_\tau)J_{n_{\tau-1},1},$$

$J_{1,k_\tau}$ is a $1 \times k_\tau$ row vector of ones,

$J_{k_\tau,1}$ is a $k_\tau \times 1$ column vector of ones.

The matrix $U_\tau$ is then defined as

$$U_\tau = \left(K_\tau^{(C)} + \sigma_\tau^2 E\right)^{-1},\qquad(6)$$

where

the matrix $K_\tau^{(C)}$ is computed from $M_\tau^{(C)}$ using (2),

$\sigma_\tau^2$ is the variance of the random variables (its role is discussed in more detail in [3]),

and $E$ is the identity matrix.

The meaning of $\sigma_\tau^2$ was analyzed in detail in [3]. In the context of the polyharmonic cascade, $\sigma_\tau^2$ can be taken to be zero (or a value close to zero if we wish to avoid a singular matrix in (6), which would correspond to an unfortunate choice of constellation, where some points are very close to each other or coincide).

If the cascade is processing a batch not only to obtain outputs, but as part of a training step, then $U_\tau$ is also used to compute an intermediate matrix $H_\tau$, which will be needed later (here $K_\tau$ is the matrix from (2)):

$$H_\tau = K_\tau U_\tau \qquad(7)$$

We now turn to the procedure for propagating derivatives: how to compute $G_{\tau-1}$ from a known matrix $G_\tau$.

First, we compute the matrix $\Theta_\tau$. It is obtained by an elementwise transformation of the matrix $M_\tau$ from (1), using

$$\theta_{ip}^\tau = \ln(m_{ip}^\tau) - 2b + 1 ,\qquad(8)$$

where

$b$ is the same coefficient as in (2),

$\theta_{ip}^\tau$ is an element of the matrix $\Theta_\tau$,

$m_{ip}^\tau$ is an element of $M_\tau$,

$i = \overline{1,r}, p = \overline{1,k_\tau}$.

We then compute the matrix $\Psi_\tau$:

$$\Psi_\tau = \Theta_\tau \circ (G_\tau \Lambda_\tau^T) \quad (9)$$

where $\circ$ denotes the Hadamard (elementwise) product.

Both $\Theta_\tau$ and $\Psi_\tau$ have size $(r \times k_\tau)$.

We can now obtain $G_{\tau-1}$:

$$G_{\tau-1} = X_{\tau-1} \circ \left(\left(\Psi_\tau J_{k_\tau,1}\right)J_{1,n_{\tau-1}}\right) - \Psi_\tau C_\tau, \quad (10)$$

where

$J_{k_\tau,1}$ is a $k_\tau \times 1$ column vector of ones,

$J_{1,n_{\tau-1}}$ is a $1 \times n_{\tau-1}$ row vector of ones.

After the matrices $H_1, H_2, \ldots, H_q$ have been computed for all polyharmonic packages in the cascade using (7), and the matrices $G_1, G_2, \ldots, G_q$ have been obtained using (10) (with the exception of $G_q$, which is not computed but simply taken to be a column vector of ones of size $(r \times 1)$), we compute the set of matrices $\Omega_1, \Omega_2, \ldots, \Omega_q$:

$$\Omega_\tau = (H_\tau H_\tau^T) \circ (G_\tau G_\tau^T) \quad (11)$$

As can be seen, (11) can be applied to each package independently (and thus computed in parallel). There is no dependence on the order of the packages in the cascade.

Next, the matrices $\Omega_1, \Omega_2, \ldots, \Omega_q$ are summed, and the vector $B$ is computed:

$$B = \left(\Omega_1 + \Omega_2 + \cdots + \Omega_q + \alpha E\right)^{-1} \Delta L \quad (12)$$

where $\Delta L = L^* - L$,

$L$ is the vector of cascade outputs,

$L^*$ is the vector of desired outputs for the cascade,

$\alpha$ is a coefficient that controls the accuracy of solving the linear system, and in effect controls the learning rate (the larger $\alpha$, the slower the learning).

The vector $B$ in (12) can be obtained either by explicitly computing the inverse matrix or by solving the corresponding linear system. Once $B$ is known, we can compute the updates to the matrices of polyharmonic function values at constellation points, $Y_\tau$, for any package:

$$\Delta Y_\tau = H_\tau^T \left(G_\tau \circ \left(B J_{1,n_\tau}\right)\right), \quad (13)$$

where $J_{1,n_\tau}$ is a $1 \times n_\tau$ row vector of ones.

The updated coefficient matrices $\Lambda_\tau$ are then obtained from (4).

Operations (13) and (4), just like (11), can be performed independently for each package (and thus computed in parallel).

Thus, expressions (1) - (13) describe both the operation of the polyharmonic cascade and the training method proposed in [5]. What remains outside this description, however, is the question of initial initialization.

If we examine what initial initialization of a polyharmonic cascade entails, it can be broken down into three components.

The first component of initialization is the structure of the cascade: how the sequence of packages is arranged, and how many inputs and outputs each package has (in the scheme above this is specified by the values $n_0, n_1, \ldots, n_{q-1}$). If the packages are connected strictly in sequence, then obviously the number of outputs of the previous package must coincide with the number of inputs of the next one. This is exactly the type of cascade considered in [5] when deriving the training algorithm. However, the resulting expressions (1) - (13) are local in nature and refer to the operations of each package separately (with the exception of (12), which, on the contrary, refers to the entire cascade as a whole). Consequently, the same training algorithm can be applied to any feedforward polyharmonic cascade of arbitrary structure. In principle, the cascade may branch into multiple paths (with different numbers of packages in each branch) and then merge again, and so on. Such variants, however, were not studied within the scope of this work.

The second component of initialization is specifying, for each package, its own constellation, which is a set of key points whose values determine all functions within that package. In the cascade scheme described in [4] and [5], constellations are represented by the matrices $C_\tau$, as if they were already given or chosen in advance. Clearly, the dimension of the space in which the constellation points are specified must coincide with the number of inputs of the package (i.e., the dimensionality of the functions computed by that package). The number of points in the constellation is not theoretically constrained and is not tied to the number of outputs of the package. In the diagram, the sizes of the constellations for the sequence of packages are denoted by $k_1, k_2, \ldots, k_q$. Accordingly, each matrix $C_\tau$ has shape $(k_\tau \times n_{\tau-1})$. Constellations are one of the key elements in the operation of polyharmonic packages. It is therefore evident that constructing the constellation matrices $C_\tau$ and assigning their values is a critically important step in the initial initialization of the cascade.

It is also clear that the constellation points should be located approximately in the same region of space as the value vectors that will appear at the input of the package – either coming from the outputs of the previous package, or, for the first package in the cascade, from the training set (in the latter case, the input values can be explicitly normalized to a suitable range).

The third component of initialization is specifying the initial values of the polyharmonic functions at the constellation points. In the scheme above, these are the initial values of the matrices $Y_\tau$. The explicit coefficients of the equations for the functions computed by the packages (the matrices $\Lambda_\tau$), as in [5], are treated as secondary, since they are fully determined by the constellation points and the function values at those points, via (4). The matrices $Y_\tau$ have shape $(k_\tau \times n_\tau)$, determined by the number of constellation points and the number of outputs of the package (i.e., the number of functions computed in that package).

Let us now examine the components of polyharmonic cascade initialization in more detail. According to the operating and training algorithm proposed in [5], the first two components, the cascade structure and the constellations, are specified at initialization and then kept fixed (during training only the matrices of function values at constellation points, $Y_\tau$, are adjusted). As numerical experiments have shown, fixing these components does not limit the ability of the cascade to learn. However, the question of whether it is beneficial to adjust not only the function values at constellation points but also the locations of the points themselves during training has not yet been investigated and remains a subject for future work.

The choice of cascade structure and the number of packages it contains can follow the same general principles as choosing the number of layers in multilayer neural networks. The more packages in the cascade and the more intermediate functions they compute, the more complex a nonlinear function the cascade as a whole can represent.

Although, in theory, even a single polyharmonic package can approximate a function of arbitrarily high complexity, this would require an unbounded increase in the number of points in its constellation. In [4], the drawbacks of using only one package, compared to cascading several packages, were analyzed. Therefore, a more practical way to increase cascade complexity (i.e., the number of trainable parameters) is to limit the constellation size in each package while increasing the number of functions computed in the packages (their outputs) and the number of packages (cascade layers).

Constellations are key elements both at initialization and during the operation of the polyharmonic cascade. There are, however, many possible ways to define them.

Below we describe one proposed method for choosing constellations – a procedure that, on the one hand, is universal and allows one to easily define a constellation for any package, and on the other hand simplifies certain operations and reduces computational cost. This procedure has been tested in practice and has demonstrated that polyharmonic cascades initialized in this way are functional and can be successfully trained.

At the same time, the algorithm proposed below is only one workable, general-purpose initialization strategy. It is quite possible that other constellation-construction schemes exist that may outperform the proposed one in some tasks or overall. The problem of designing constellations is itself a promising direction for further research.

### Constructing constellations: proposed scheme

What properties should we expect from a universal procedure for constructing constellations? First, it should be applicable in spaces of arbitrary dimension (in accordance with the number of inputs to a package). Second, one may reasonably hypothesize that, since we do not know in advance which functions a given package will need to compute, a good default is to place the constellation points in some symmetric way in this multidimensional space. Under these considerations, natural candidates for constellations (and those that should be tested first) are point sets forming the vertices of regular multidimensional polytopes. One of the relatively simple options, both to construct and to use, is the hyperoctahedron, although polytopes of other types may also be viable candidates.

In the present work, as part of the numerical experiments, an algorithm for automatic construction of constellations at cascade initialization was implemented as follows: the first constellation point is placed at the origin, and then the vertices of a hyperoctahedron are placed around it at unit distance.

Figure 2 illustrates examples of such constellations in spaces of dimension one through four. By the same principle, one can define a constellation in a space of arbitrary dimension. If the dimensionality of the space is $n$, then the constellation will contain $2n + 1$ points.

In the scheme considered earlier (Figure 1), the number of inputs of the package with index $\tau$ was denoted by $n_{\tau-1}$. Hence, under the proposed scheme, the number of points in its constellation, denoted $k_\tau$, is

$$k_\tau = 2n_{\tau-1} + 1 \qquad (14)$$

In what follows, however, we will consider formulas referring to a single, generic package with index $\tau$ (and thus applicable to any package). To avoid cluttering the notation, we will simply denote the number of inputs of this package by $n$, implicitly assuming $n = n_{\tau-1}$.

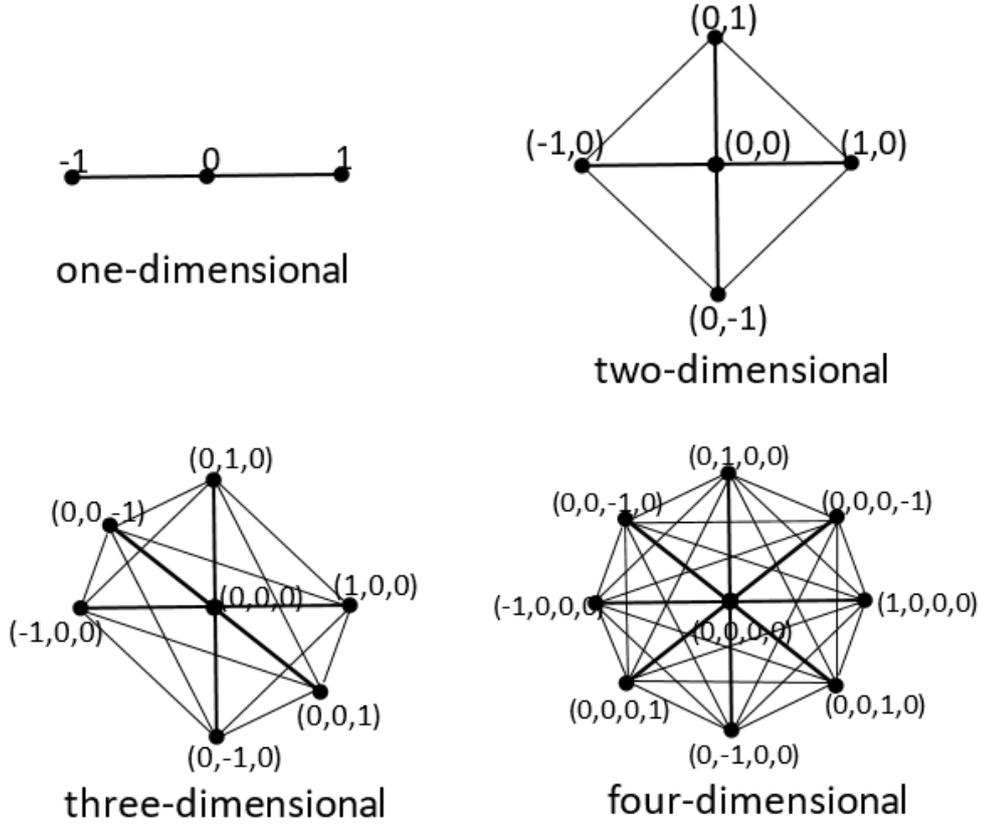

Figure 2. Principles of constructing constellations based on hyperoctahedra.

We now describe how such a constellation is represented in matrix form as $C_\tau$. For a hyperoctahedral constellation in $n$ dimensions we have

$$C_\tau = \begin{bmatrix} 0_{1 \times n} \\ -E_n \\ E_n \end{bmatrix}, \qquad (15)$$

where
$E_n$ is the $(n \times n)$ identity matrix,
$0_{1 \times n}$ is a row vector of zeros of length $n$.

Evidently, for a polyharmonic cascade with constellations defined in this way to operate properly, the values entering each package should lie not too far from the interval $(-1,1)$ along each coordinate. For the first package in the cascade, this requirement can be satisfied by normalizing the input data. One can also choose initial values of the polyharmonic functions at the constellation points (the matrices $Y_\tau$) so that the cascade is already functional at the very beginning of training.

A natural concern is whether, during training, the outputs of some packages might drift far outside the region covered by the fixed constellations. The training algorithm from [5] does not explicitly constrain the outputs of the packages in any way. However, numerical experiments show that, with a suitable choice of the parameter $\alpha$ in (12) and with the initialization procedure described in this paper, the polyharmonic cascade behaves stably regardless of the number of packages. Training according to the algorithm in [5] forces the trainable polyharmonic functions in different packages to act coherently, as a single whole.

Let us now examine in more detail how the choice of a hyperoctahedral constellation of the form (15) affects the operations (1) - (13).

First, expression (1) simplifies significantly if we substitute (15) for $C_\tau$:

$$M_\tau = N_{\tau x} J_{1,k_\tau} + [0_{r\times 1} \quad J_{r,k_\tau} + 2X_{\tau-1} \quad J_{r,k_\tau} - 2X_{\tau-1}], \qquad (16)$$

where

$$N_{\tau x} = (X_{\tau-1} \circ X_{\tau-1}) J_{n,1},$$

$0_{r\times 1}$ is an $r \times 1$ column vector of zeros,
$J_{1,k_\tau}$ is a $1 \times k_\tau$ row vector of ones,
$J_{n,1}$ is an $n \times 1$ column vector of ones,
$J_{r,k_\tau}$ is an $r \times k_\tau$ matrix of ones.

In (1), the most computationally expensive term was the matrix product $X_{\tau-1} C_\tau^T$. As we can see, when using the constellation (15), expression (1), rewritten as (16), becomes radically simpler.

Expression (5) also undergoes substantial simplification. For the hyperoctahedral constellation (15), we have

$$M_\tau^{(C)} = N_{\tau c} J_{1,k_\tau} + J_{k_\tau,1} N_{\tau c}^T - 2C_\tau C_\tau^T =$$

$$= \begin{bmatrix} 0_{1\times 2n+1} \\ J_{2n,2n+1} \end{bmatrix} + [0_{2n+1\times 1} \quad J_{2n+1,2n}] - 2\begin{bmatrix} 0 & 0_{1\times n} & 0_{1\times n} \\ 0_{n\times 1} & E_n & -E_n \\ 0_{n\times 1} & -E_n & E_n \end{bmatrix}, \qquad (17)$$

where

$0_{1\times 2n+1}$ is a row of zeros of length $2n+1$,
$J_{2n,2n+1}$ is a $2n \times (2n+1)$ matrix of ones,
$0_{2n+1\times 1}$ is a $(2n+1) \times 1$ column of zeros,
$J_{2n+1,2n}$ is a $(2n+1) \times 2n$ matrix of ones,
$0_{1\times n}$ is a row of zeros of length $n$,
$0_{n\times 1}$ is an $n \times 1$ column of zeros.

Introduce the matrix

$$P_{2n} = \begin{bmatrix} 0_{n \times n} & E_n \\ E_n & 0_{n \times n} \end{bmatrix} \quad (18)$$

Then (17) can be rewritten as

$$M_\tau^{(C)} = \begin{bmatrix} 0 & J_{1,2n} \\ J_{2n,1} & 2J_{2n} \end{bmatrix} - 2 \begin{bmatrix} 0 & 0_{1 \times 2n} \\ 0_{2n \times 1} & E_{2n} - P_{2n} \end{bmatrix}, \quad (19)$$

where $E_{2n}$ is the $2n \times 2n$ identity matrix,

and $J_{2n}$ is the $2n \times 2n$ matrix of ones.

Finally, we obtain

$$M_\tau^{(C)} = \begin{bmatrix} 0 & J_{1,2n} \\ J_{2n,1} & 2J_{2n,2n} - 2E_{2n} + 2P_{2n} \end{bmatrix} \quad (20)$$

Let us examine the structure of (20) in more detail.

Along the main diagonal (i.e., in the first element of the first row and in those elements where $E_{2n}$ in (20) equals 1), the entries of $M_\tau^{(C)}$ are equal to 0. In the first row and the first column of (20), except for the first diagonal element, $M_\tau^{(C)}$ takes the value 1.

In those positions where $P_{2n}$ in (20) has entry 1, $M_\tau^{(C)}$ takes the value 4.

All the remaining entries of $M_\tau^{(C)}$ are equal to 2.

Thus, the entries of the matrix in (20) can take only four possible values: 0, 1, 2, or 4. We can exploit this when computing $K_\tau^{(C)}$ (which is obtained from $M_\tau^{(C)}$ using (2)). Denote by $k_0, k_1, k_2, k_4$ the values of the scalar function (2) at inputs 0, 1, 2, and 4, respectively:

$$k_0 = \lim_{x \to 0}(0.5x(\ln(x) - 2b) + c) = c \quad (21)$$

$$k_1 = 0.5(\ln(1) - 2b) + c = c - b \quad (22)$$

$$k_2 = (\ln(2) - 2b) + c \quad (23)$$

$$k_4 = 2(\ln(4) - 2b) + c \quad (24)$$

Using (20) - (24), we can now explicitly write $K_\tau^{(C)}$:

$$K_\tau^{(C)} = \begin{bmatrix} k_0 & k_1 J_{1,2n} \\ k_1 J_{2n,1} & (k_0 - k_2)E_{2n} + (k_4 - k_2)P_{2n} + k_2 J_{2n} \end{bmatrix}, \quad (25)$$

where $J_{1,2n}$ and $J_{2n,1}$ are a $1 \times 2n$ row and a $2n \times 1$ column of ones, respectively, and $J_{2n}$ is the $2n \times 2n$ all-ones matrix.

Using (25), we can now compute $U_\tau$ as in (6). First, let us write out the matrix $K_\tau^{(C)} + \sigma_\tau^2 E$, whose inverse we need:

$$K_\tau^{(C)} + \sigma_\tau^2 E = \begin{bmatrix} k_0 + \sigma_\tau^2 & k_1 J_{1,2n} \\ k_1 J_{2n,1} & (k_0 - k_2 + \sigma_\tau^2) E_{2n} + (k_4 - k_2) P_{2n} + k_2 J_{2n} \end{bmatrix} \quad (26)$$

For the hyperoctahedral constellation (15), we can safely take $\sigma_\tau^2 = 0$.

To compute $U_\tau$ we must invert the block matrix in (26). Since it is a $2 \times 2$ block matrix, we use the Frobenius formula (block inversion formula):

$$\begin{bmatrix} K_1 & K_2 \\ K_3 & K_4 \end{bmatrix}^{-1} = \begin{bmatrix} K_1^{-1} + K_1^{-1} K_2 S^{-1} K_3 K_1^{-1} & -K_1^{-1} K_2 S^{-1} \\ -S^{-1} K_3 K_1^{-1} & S^{-1} \end{bmatrix},$$

where $S = K_4 - K_3 K_1^{-1} K_2$

For the matrix in (26) the blocks are:

$$K_1 = k_0 + \sigma_\tau^2$$
$$K_2 = k_1 J_{1,2n}$$
$$K_3 = k_1 J_{2n,1}$$

$$S = (k_0 - k_2 + \sigma_\tau^2) E_{2n} + (k_4 - k_2) P_{2n} + k_2 J_{2n} - k_1 J_{2n,1} \frac{1}{k_0 + \sigma_\tau^2} k_1 J_{1,2n} =$$

$$= (k_0 - k_2 + \sigma_\tau^2) E_{2n} + (k_4 - k_2) P_{2n} + \left(k_2 - \frac{k_1^2}{k_0 + \sigma_\tau^2}\right) J_{2n} \quad (27)$$

Introduce the coefficients

$$a_1 = k_0 - k_2 + \sigma_\tau^2 \quad (28)$$
$$a_2 = k_4 - k_2 \quad (29)$$
$$a_3 = k_2 - \frac{k_1^2}{k_0 + \sigma_\tau^2} \quad (30)$$

Then we can write

$$S = a_1 E_{2n} + a_2 P_{2n} + a_3 J_{2n} \quad (31)$$

By the Frobenius formula, $U_\tau$ takes the form

$$U_\tau = \begin{bmatrix} \frac{1}{k_0 + \sigma_\tau^2} + \frac{k_1^2}{(k_0 + \sigma_\tau^2)^2} J_{1,2n} S^{-1} J_{2n,1} & -\frac{k_1}{k_0 + \sigma_\tau^2} J_{1,2n} S^{-1} \\ -\frac{k_1}{k_0 + \sigma_\tau^2} S^{-1} J_{2n,1} & S^{-1} \end{bmatrix} \quad (32)$$

Let us represent $S^{-1}$ in the same basis as

$$S^{-1} = b_1 E_{2n} + b_2 P_{2n} + b_3 J_{2n} \tag{33}$$

The product of (31) and (33) must equal the identity matrix $E_{2n}$ (of size $2n \times 2n$):

$$E_{2n} = (a_1 E_{2n} + a_2 P_{2n} + a_3 J_{2n})(b_1 E_{2n} + b_2 P_{2n} + b_3 J_{2n}) =$$
$$= a_1 b_1 E_{2n} + a_1 b_2 P_{2n} + a_1 b_3 J_{2n} + a_2 b_1 P_{2n} + a_2 b_2 E_{2n} + a_2 b_3 J_{2n} +$$
$$+ a_3 b_1 J_{2n} + a_3 b_2 J_{2n} + a_3 b_3 2n J_{2n} = (a_1 b_1 + a_2 b_2) E_{2n} + (a_1 b_2 + a_2 b_1) P_{2n} +$$
$$+ (a_1 b_3 + a_2 b_3 + a_3 b_1 + a_3 b_2 + 2n a_3 b_3) J_{2n} \tag{34}$$

From (34) we obtain the system of equations

$$\begin{cases} a_1 b_1 + a_2 b_2 = 1 \\ a_1 b_2 + a_2 b_1 = 0 \\ a_1 b_3 + a_2 b_3 + a_3 b_1 + a_3 b_2 + 2n a_3 b_3 = 0 \end{cases} \tag{35}$$

Solving this system gives

$$b_1 = \frac{a_1}{a_1^2 - a_2^2} \tag{36}$$

$$b_2 = -\frac{a_2}{a_1^2 - a_2^2} \tag{37}$$

$$b_3 = -\frac{a_3}{(a_1 + a_2 + 2n a_3)(a_1 + a_2)} \tag{38}$$

Since $S^{-1}$ is expressed in the form (33) – as a weighted sum of the identity matrix, the matrix $P_{2n}$ from (18), and the all-ones matrix – the sum of all elements in any row or column of $S^{-1}$ equals $b_1 + b_2 + 2n b_3$. The sum of all elements of $S^{-1}$ is therefore $2n(b_1 + b_2 + 2n b_3)$.

We can exploit these properties when substituting $S^{-1}$ into (32).

Introduce the additional coefficients $u_1$ and $u_2$:

$$u_1 = \frac{1}{k_0 + \sigma_\tau^2} + \frac{k_1^2}{(k_0 + \sigma_\tau^2)^2} J_{1,2n} S^{-1} J_{2n,1} =$$

$$= \frac{1}{k_0 + \sigma_\tau^2} + \frac{k_1^2}{(k_0 + \sigma_\tau^2)^2} 2n(b_1 + b_2 + 2n b_3) \tag{39}$$

$$u_2 = -\frac{k_1}{k_0 + \sigma_\tau^2}(b_1 + b_2 + 2n b_3) \tag{40}$$

Then, using (32) - (33) and (36) - (40), we can write $U_\tau$ in compact form:

$$U_\tau = \begin{bmatrix} u_1 & u_2 J_{1,2n} \\ u_2 J_{2n,1} & b_1 E_{2n} + b_2 P_{2n} + b_3 J_{2n} \end{bmatrix} \quad (41)$$

As can be seen from expression (41), when using constellations of the hyperoctahedral form (15), there is no need to compute either the matrix $M_\tau^{(C)}$ or the matrix $K_\tau^{(C)}$ explicitly, nor to invert the matrix in (6) directly in order to obtain $U_\tau$. Instead, one can first compute $k_0, k_1, k_2, k_4$ from (21) - (24), then obtain the coefficients $a_1, a_2, a_3$ from (28) - (30), which in turn yield $b_1, b_2, b_3$ via (36) - (38), and finally compute $u_1$ and $u_2$ from (39) - (40). This is sufficient to synthesize $U_\tau$ through (41). As a result, the amount of computation is drastically reduced and becomes independent of the constellation size (15), and likewise independent of the number of inputs of the package (i.e., the dimensionality of the functions approximated by the package).

We can, however, go one step further and look at what $U_\tau$ is actually used for.

The matrix $U_\tau$ serves to transform the values of the polyharmonic functions $Y_\tau$ at the constellation points into the coefficients $\Lambda_\tau$ of the polyharmonic functions themselves (needed for their evaluation), as in (4). In addition, $U_\tau$ is used during training to form the matrix $H_\tau$ in (7), which is then used in (11).

Represent the matrix $Y_\tau$ in the form

$$Y_\tau = \begin{bmatrix} y_1 \\ Y_1 \\ Y_2 \end{bmatrix}, \quad (42)$$

where

$y_1$ is the first row of $Y_\tau$;

$Y_1$ and $Y_2$ are matrices that together collect the next $n$ rows each of $Y_\tau$.

Define

$$Y_{1+2} = \begin{bmatrix} Y_1 \\ Y_2 \end{bmatrix}, \quad Y_{2+1} = \begin{bmatrix} Y_2 \\ Y_1 \end{bmatrix}, \quad y_s = J_{1,2n} Y_{1+2} \quad (43)$$

where $Y_{1+2}$ and $Y_{2+1}$ contain the same blocks $Y_1$ and $Y_2$, but in different order, and the row vector $y_s$ is the sum of the columns of $Y_{1+2}$.

Now combine (4), (41), (42), and (43):

$$\Lambda_\tau = U_\tau Y_\tau = \begin{bmatrix} u_1 y_1 + u_2 J_{1,2n} Y_{1+2} \\ u_2 J_{2n,1} y_1 + (b_1 E_{2n} + b_2 P_{2n} + b_3 J_{2n}) Y_{1+2} \end{bmatrix} =$$

$$= \begin{bmatrix} u_1 y_1 + u_2 y_s \\ b_1 Y_{1+2} + b_2 Y_{2+1} + J_{2n,1}(u_2 y_1 + b_3 y_s) \end{bmatrix} \quad (44)$$

Next, write the matrix $K_\tau$, obtained from $M_\tau$ via (2) (with $M_\tau$ itself given by (2) or now by (16)), in the form

$$K_\tau = [\hat{k}_1 \quad K_1 \quad K_2] \qquad (45)$$

where

$\hat{k}_1$ is the first column vector of $K_\tau$;

$K_1$ and $K_2$ are matrices that together collect the next $n$ columns each of $K_\tau$.

Define

$$K_{1+2} = [K_1 \quad K_2], \quad K_{2+1} = [K_2 \quad K_1], \quad \hat{k}_s = K_{1+2} J_{2n,1} \qquad (46)$$

Here $K_{1+2}$ and $K_{2+1}$ contain the same blocks $K_1$ and $K_2$ in different order, and the column vector $\hat{k}_s$ is the sum of the rows of $K_{1+2}$.

Combining (7), (41), (45), and (46) gives

$$H_\tau = K_\tau U_\tau = [u_1 \hat{k}_1 + u_2 \hat{k}_s \quad b_1 K_{1+2} + b_2 K_{2+1} + (u_2 \hat{k}_1 + b_3 \hat{k}_s) J_{1,2n}] \qquad (47)$$

Thus, using (44) and (47), we can compute $\Lambda_\tau$ and $H_\tau$ directly from $Y_\tau$ and $K_\tau$. Compared to (4) and (7), the computational complexity is significantly reduced. The only prerequisite is to precompute, once at cascade initialization, the coefficients $b_1, b_2, b_3, u_1, u_2$ for each package. There is no longer any need to compute and store the matrices $U_\tau$ explicitly (nor $M_\tau^{(C)}$ and $K_\tau^{(C)}$).

Finally, consider expression (10). Write $\Psi_\tau$ as

$$\Psi_\tau = [\psi_1 \quad \Psi_1 \quad \Psi_2] \qquad (48)$$

where

$\psi_1$ is the first column of $\Psi_\tau$;

$\Psi_1$ and $\Psi_2$ are matrices that together collect the next $n$ columns each of $\Psi_\tau$.

Then, substituting (15) into (10), we can rewrite it as

$$G_{\tau-1} = X_{\tau-1} \circ \left((\Psi_\tau J_{k_\tau,1}) J_{1,n_{\tau-1}}\right) + [\Psi_1 - \Psi_2] \qquad (49)$$

where $\circ$ denotes the Hadamard (elementwise) product.

As a result, we eliminate from (10) the most computationally expensive operation – the matrix multiplication with the explicit constellation matrix.

Taken together, if we now analyze expressions (14) - (49), we can conclude that using the constellation construction based on hyperoctahedra (plus the central point) provides not only a universal way to define constellations for any package, but also allows one to significantly simplify the computations. There is no need to store and use the constellation matrices $C_\tau$ explicitly. Instead, one can work with expressions (16) and (49), into which (1) and (10) are transformed.

Likewise, at initialization there is no need to compute and store $U_\tau$ in order to apply (4) and (7). Instead, one can compute once (at cascade initialization) the coefficients $b_1, b_2, b_3, u_1, u_2$ using (21)–(24), (28)–(30), and (36)–(40), and then replace (4) and (7) by (44) and (47), respectively.

Thus, using hyperoctahedra as constellations can yield a substantial computational advantage. At the same time, it is reasonable to expect that other symmetric high-dimensional figures used as constellations may also lead to computational optimizations, but would result in alternative formulas instead of (16), (44), (47), and (49). As an alternative for more uniform coverage of the space, one could consider LHS/quasi–Monte Carlo constructions or spherical designs (McKay et al. [15], Niederreiter [17]; Delsarte et al. [8]). This is a natural direction for further research.

It is worth noting that in the computational experiments, in the PyTorch implementation (though this may partly reflect specifics of that particular implementation), the formulas (16), (44), (47), and (49), despite requiring far fewer arithmetic operations, start to show a clear speed advantage over (1), (4), (7), and (10) only when the number of inputs in a package (and hence the matrix sizes) exceeds several hundred (this was observed both on CPU and GPU). This is likely due to the much heavier relative optimization of matrix multiplication, both in hardware and software, compared to other, simpler operations. For example, splitting the matrix $\Psi_\tau$ into blocks as in (48) and performing the subtraction in (49) may run slower than multiplying $\Psi_\tau$ by $C_\tau$ in (10), if their dimensions are below a few hundred rows. At the same time, this observation suggests that, in principle, the polyharmonic cascade can be implemented more efficiently in the future than in the current codebase used for the experiments whose results are reported below.

Let us now turn to another important aspect of initializing the polyharmonic cascade, namely the initial values at the constellation points, i.e., the initial values of the matrices $Y_\tau$. Their choice determines which functions each package will compute at the very beginning, layer by layer, when the cascade is first constructed.

One of the key steps in training the polyharmonic cascade is the sequential computation (from the last package backward to the first) of the partial derivative matrices $G_\tau$. Although these derivatives are not used directly for optimization in the manner of gradient descent, they are crucial for forming the systems of linear equations.

If, for example, in some package we assign the same fixed value (not necessarily zero) to all entries of the matrix $Y_\tau$, then all functions at the output of this package will always return this fixed value (or something very close to it), and their derivatives will be zero everywhere (or very close to zero). As a result, for all packages that precede package $\tau$, all derivative matrices will collapse to zero, and those packages will no longer be able to learn (i.e., change their own matrices $Y_\tau$), which follows directly from (13). In practice this would only cause trouble on the very first iteration, after which the values in $Y_\tau$ may change, and on the next iteration the entire cascade could begin to learn. However, if such an unfortunate choice of $Y_\tau$ (all entries nearly identical) is made for two or more packages in the cascade, this will effectively render the cascade non-functional.

From these considerations it follows that each $Y_\tau$ must consist of sufficiently different elements. As initial $Y_\tau$, one should not use zero matrices or matrices filled with the same constant value in more than one package.

What range should the initial values in $Y_\tau$ lie in? Clearly, they should lie in the same range as the constellation points in the next package. Since we have chosen constellations to be hyperoctahedra whose vertices lie at unit distance from the origin, it is natural to take the initial entries of $Y_\tau$ in the range approximately from $-1$ to $1$, and at the same time ensure that the lengths of the row vectors of $Y_\tau$ do not deviate too far from one.

Based on these considerations, one can propose the following initialization procedure for $Y_\tau$. The matrices $Y_\tau$ may be filled with random values in the interval $[-1,1]$, subject to a constraint on the norms of the resulting row vectors (for example, one may first fill them with random values and then normalize each row to have approximately unit length).

Why not, then, simply choose the values at the constellation points $Y_\tau$ so that the outputs of a given package coincide exactly with the constellation of the next package? In principle, this is entirely possible. However, with the initialization procedure based on hyperoctahedral constellations described in this paper, this can only be done when successive packages have the same number of inputs and outputs. Even so, this idea can be exploited, at least for certain fragments of the polyharmonic cascade rather than for the entire cascade.

Assume that in the polyharmonic cascade there is a fragment consisting of a sequence of packages, from index $\tau$ to index $t$, such that each of them has the same number of inputs and outputs:

$$n_{\tau-1} = n_\tau = n_{\tau+1} = \cdots = n_{t-1} = n_t = n$$

Then, if we use the constellation initialization procedure proposed in this paper, all these packages will have identical constellations with the same number of points:

$$k_\tau = k_{\tau+1} = \cdots k_{t-1} = k_t = 2n + 1 = k$$

For brevity, denote the number of inputs (and outputs) in each of these packages by $n$, and the number of constellation points in each of them by $k$. In this case, we obtain the structure illustrated in Figure 3.

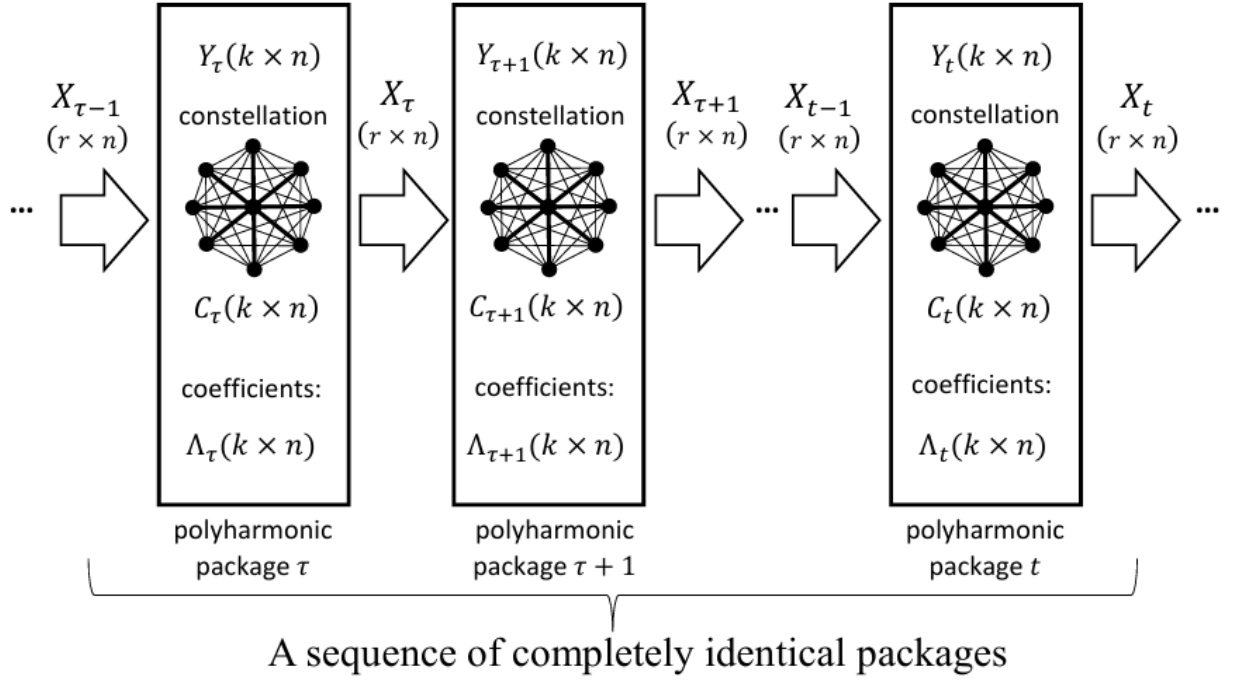

Figure 3. A fragment of the polyharmonic cascade

In such a situation, during initialization of this sequence of packages one can simply set the matrices $Y_\tau$ equal to the corresponding constellation matrices $C_\tau$:

$$Y_\tau = Y_{\tau+1} = \cdots = Y_t = C_\tau = \cdots = C_t \qquad (50)$$

If we choose $\sigma_\tau^2$ from (6) or (28), (30) to be zero, and if the row vectors of $X_{\tau-1}$ lie within the region covered by the constellation of package $\tau$, then the matrix $X_{\tau-1}$, after passing through all these packages (in the state they have immediately after initialization), will propagate to $X_t$ almost unchanged, i.e.,

$$X_t \approx X_{\tau-1}.$$

Exploiting this property, one can, if necessary, increase the number of packages in the cascade essentially without limit, while preserving its operability and its ability to learn after initialization.

Finally, let us briefly discuss some additional possibilities for optimizing computations during training. Consider the structure of the matrix $K_\tau$ from (3) for the first package in the cascade, i.e., the matrix $K_1$. Each of its rows is determined entirely by the corresponding input vector from the batch. In the next epoch, this vector will appear in a different batch, but the corresponding row in $K_1$ will be exactly the same. Therefore, one can precompute the entire input part of the training set and obtain all possible rows of the matrix $K_1$ in advance. Training batches can then be fed directly in the form of ready-made $K_1$ matrices into (3) or (47), bypassing (16) and (2). One can go further and apply the same idea directly to the matrix $H_1$ from (7) or (47). In that case, $K_1$ is no longer needed during training at all, and (3) can be replaced by

$$X_1 = H_1 Y_1 \tag{51}$$

Additional computational savings can also be achieved when computing the vector $B$ in (12) by solving the corresponding linear system. The matrix $(\Omega_1 + \Omega_2 + \cdots + \Omega_q + \alpha E)$ in (12) is always symmetric and positive definite. Consequently, one can use the Cholesky decomposition to solve the system efficiently – this is a standard tool in numerical linear algebra for symmetric positive definite matrices (Golub & Van Loan, 2013 [11]).

Thus, taking into account the theoretical foundations laid out in [3], [4], and [5], together with the initialization principles for the polyharmonic cascade proposed in this paper, we have now described all the components needed to implement this machine learning architecture in code, to launch it, and to verify its operability.

**Launch and validation.**

Below we present the results of the computational experiments. All experiments were run on a machine with the following configuration: Intel Core i9-10920X CPU and an NVIDIA GeForce RTX 3070 GPU with 8 GB of memory. All computations related to the polyharmonic cascade were executed on the GPU.

The parameters $b$ and $c$ from (2) (and hence from (21) - (24)) were fixed to $b$=5, $c$=400. In all experiments, arithmetic was performed in float32.

The source code required to reproduce the experiments is available in a public GitHub repository: https://github.com/xolod7/polyharmonic-cascade.git [22], and is archived on Zenodo with the permanent DOI: https://doi.org/10.5281/zenodo.16811633 [23].

MNIST

For the first series of experiments we used the MNIST dataset ([14], loaded via the torchvision library). The training set contains 60,000 examples and the test set 10,000 examples.

The data were used in their original form, without any preprocessing, augmentation, or extraction of image features. During training, we did not exploit the fact that the inputs are images: the cascade simply received 784-dimensional vectors as input ("a bag of pixels").

Given this setup, the goal of the MNIST experiments was not to achieve results close to state-of-the-art (SOTA) models on this benchmark. Rather, the aim was to verify the basic operability of the polyharmonic cascade as a machine learning architecture, and its ability to train using the same learning algorithm [5] regardless of cascade configuration (including extreme ones), even on a relatively small dataset. In other words, the goal was to test the scalability of the polyharmonic cascade and the robustness of the underlying algorithms for arbitrary cascade configurations, with MNIST serving as a convenient working example.

Scaling to 10 outputs was carried out using the procedure described in [5], by replicating the cascade structure for each output. Under these conditions, validating the fundamental operability of the polyharmonic cascade on MNIST is in fact a stricter test than predicting a single scalar output: it suffices for just one of the ten classifiers to train poorly for the overall result to degrade. The batch size in all MNIST experiments was set to 2000.

Experiment (MNIST) 1.

In this experiment we used a polyharmonic cascade with architecture 784–100–20–20–10, i.e., four packages. The total number of trainable parameters in the cascade was about 1.6 million. Training was carried out for 10 epochs. The value of $\alpha$ in (12) was set to 200.

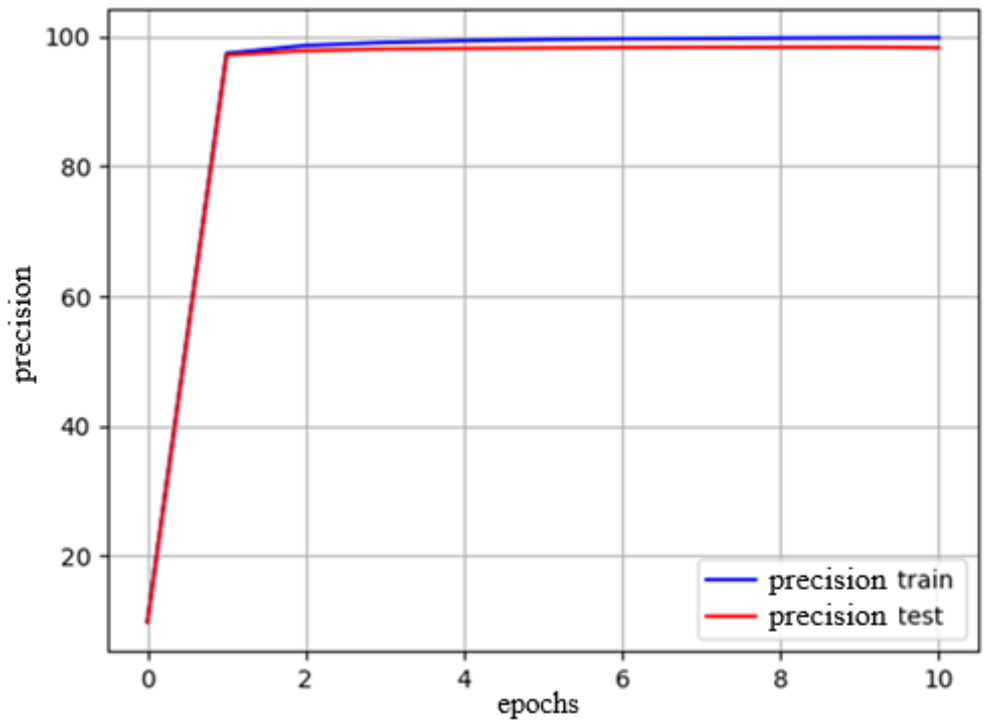

Figure 4. Experiment 1. Training curve on the MNIST dataset.

The time per epoch was approximately 1.4-1.5 seconds. After the first epoch, the test accuracy reached 97.14%.

If we zoom in on the upper part of the curve:

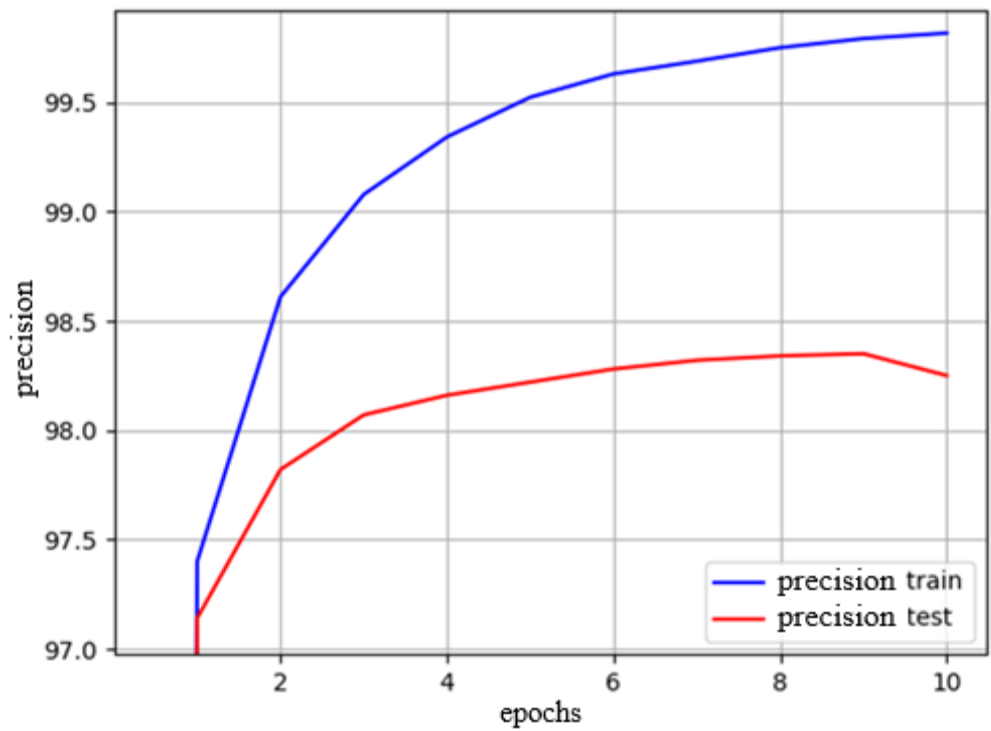

Figure 5. Experiment 1. Training curve on the MNIST dataset (zoomed).

Over 10 epochs, the test accuracy reached 98.35%. On different runs this value varies slightly, but it is consistently above 98.1%.

Thus, the first experiment confirmed the applicability of the polyharmonic cascade as a machine learning architecture, as well as the effectiveness of the training method proposed in [5] and the hyperoctahedron-based initialization procedure introduced in this paper. The achieved accuracy of 98.35% is comparable to that of other methods reported for MNIST (see [14]), under the constraint that we did not exploit, either explicitly or implicitly through the architecture (as in convolutional networks), the fact that the data are images with spatially organized pixels.

Experiment (MNIST) 2.

To test the scalability of the algorithms, we increased the number of trainable parameters in the cascade to 75.74 million (almost a 50-fold increase compared to Experiment 1). This corresponds to a cascade of 5 layers with architecture 784–1000–1000–1000–1000–10. The value of $\alpha$ was kept the same as before, $\alpha=200$.

Training was run for 10 epochs.

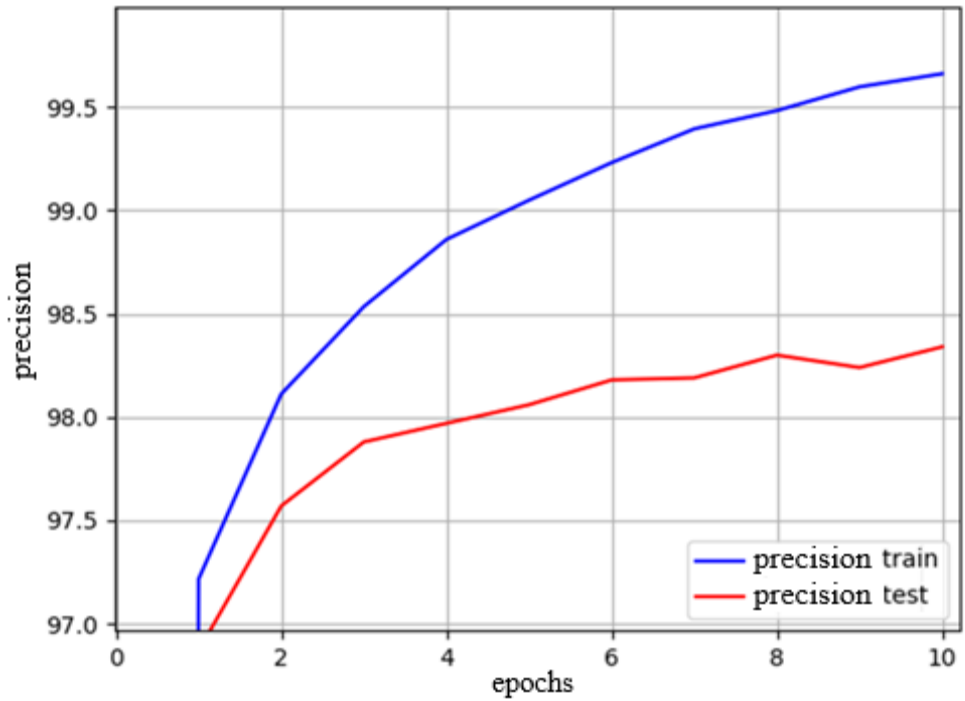

Figure 6. Experiment 2 (MNIST). Training over 10 epochs.

The time per epoch increased to 12–17 seconds (about a 10-fold increase compared to Experiment 1, even though the number of parameters grew by nearly a factor of 50). The evolution of test accuracy was qualitatively similar to that observed in Experiment 1.

Next, we examined the learning dynamics when the number of epochs was increased to 100.

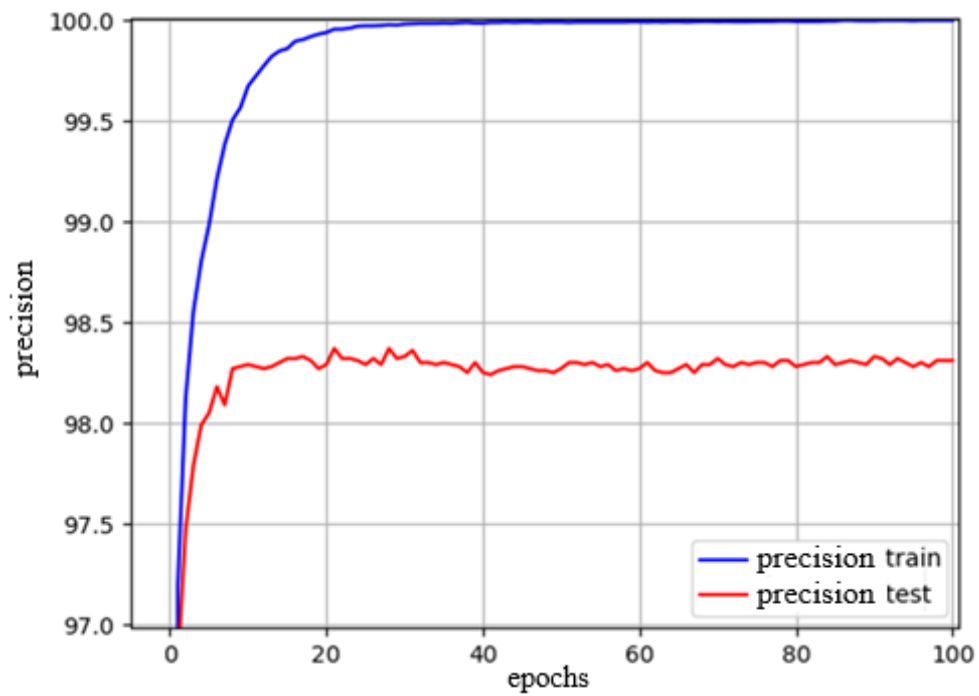

Figure 7. Experiment 2 (MNIST). Training over 100 epochs.

The training accuracy reached nearly 100%. Test accuracy stabilized in the range 98.25 - 98.3%. No pronounced overfitting was observed as training proceeded for this task.

It is plausible that a cascade with 75 million parameters is excessive for MNIST and partially "memorizes" some training examples instead of generalizing. To investigate this, we deliberately slowed down learning by setting α=10000 and trained for 300 epochs.

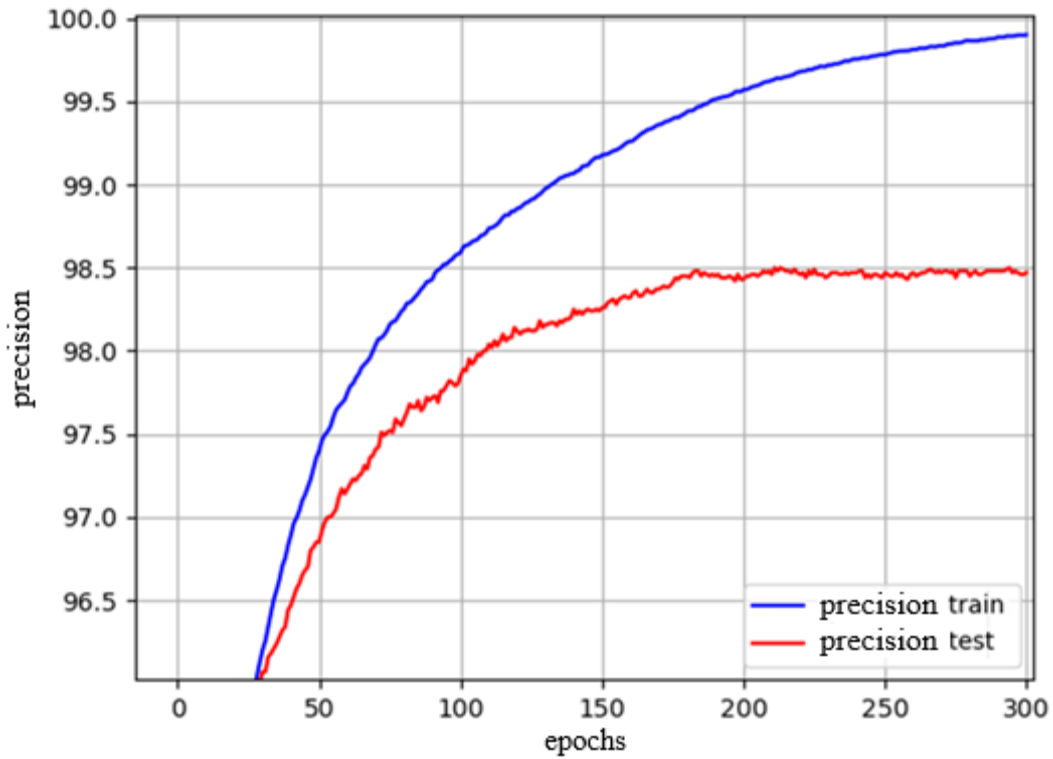

Figure 8. Experiment 2 (MNIST). Intentionally slowed training.

As can be seen, this had a beneficial effect on performance: the cascade surpassed 98.45% test accuracy.

Experiment (MNIST) 3.

We now test the scalability of the polyharmonic cascade and its training algorithm in a different way. Instead of increasing the number of inputs and outputs in each package, we significantly increase the number of packages in the cascade.

Consider a cascade consisting of 100 layers with architecture 784–100–100–…–100–100–10.

All intermediate packages (i.e., all except the first and the last) have 100 inputs and 100 outputs. In this case, we can use the initialization (50). The resulting cascade contains 21.27 million trainable parameters. The parameter $\alpha$ was set to 50.

We trained this model for 10 epochs.

The time per epoch was approximately 25.5 seconds. After the first epoch, the test accuracy was about 97%. After a few more epochs, the accuracy stabilized around 98%.

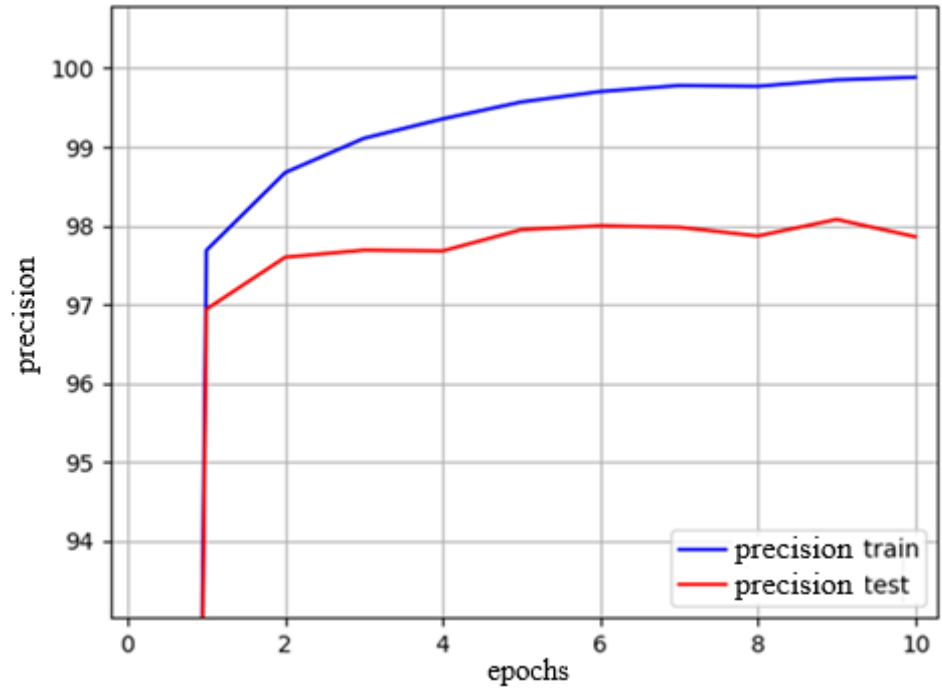

Figure 9. Experiment 3 (MNIST). Training a 100-layer cascade.

A cascade with 100 layers is probably not the most optimal model for MNIST classification, but the result demonstrates that as we increase the number of packages in the polyharmonic cascade, it remains functional, can be initialized using the procedure proposed in this paper, and can be trained with the same algorithm without any fundamental modifications (aside from possibly choosing a more optimal value of $\alpha$).

Experiment (MNIST) 4.

We now consider a cascade consisting of 500 layers with architecture 784–100–25–25–…–25–10. This yields a cascade with an even larger number of packages than before, but most of them have only 25 inputs and 25 outputs.

The resulting cascade contains 7.95 million trainable parameters. We set $\alpha=2000$ and trained for 10 epochs. As can be seen from the corresponding figure, the cascade exhibited qualitatively the same accuracy dynamics as in Experiment 1, achieving a substantial improvement already in the first epoch.

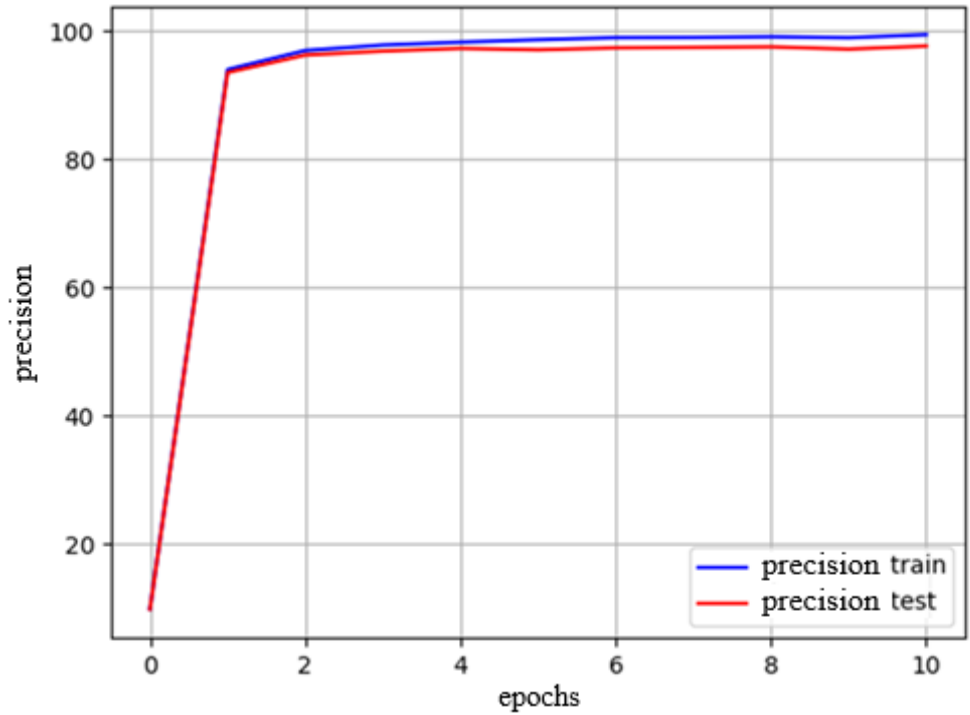

Figure 10. Experiment 4 (MNIST). Training a 500-layer cascade.

Let us look more closely at the accuracy curves:

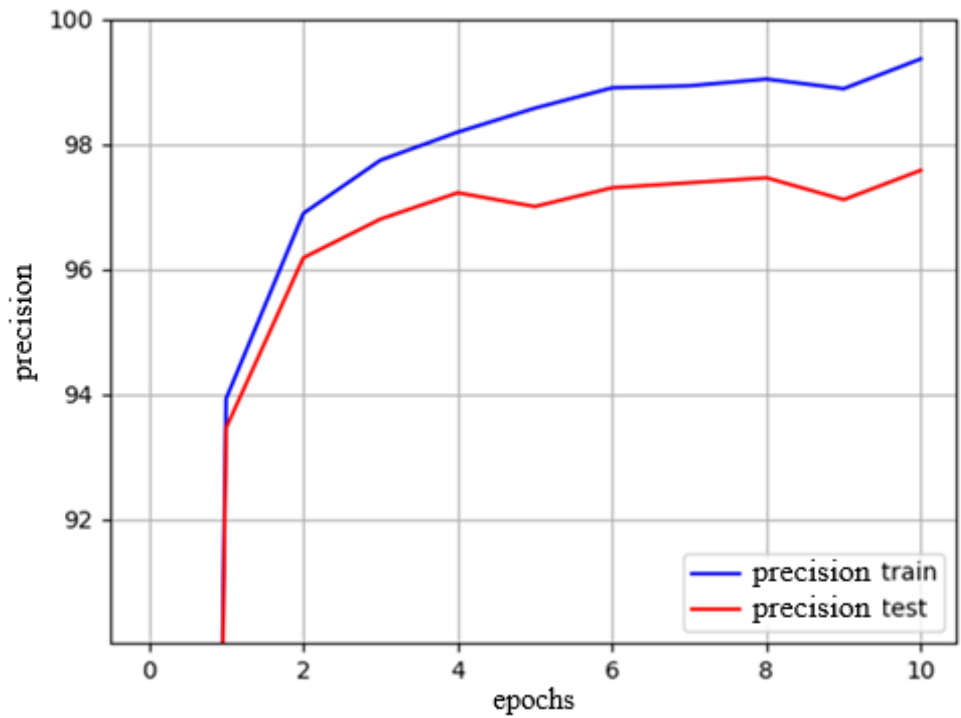

Figure 11. Experiment 4 (MNIST). Training a 500-layer cascade (zoomed).

The final accuracies are lower than in the previous experiments; however, the results show that even a 500-layer cascade remains functional and can be trained with exactly the same algorithm as a 4-layer cascade. The packages are connected strictly in sequence, without any additional mechanisms such as skip connections that are commonly used to build very deep neural network architectures. It is plausible that using skip connections (or similar techniques) could further improve the performance of the polyharmonic cascade, but this was not investigated in the present work and is left for future research.

It is also worth noting that in this fourth experiment the time per epoch increased to about 70 seconds. Despite the fact that this cascade has fewer parameters than in Experiment 2, each epoch is significantly slower. This is likely due to GPU execution characteristics: multiplications of larger matrices are often more efficient in terms of the number of elementary arithmetic operations per unit time than many smaller matrix operations.

Interim conclusion.

The MNIST experiments confirm the operability of polyharmonic cascades across a wide range of sizes and depths. The same training algorithm can be applied both to a 4-layer and to a 500-layer cascade, and likewise to models with 1.6 million and with 75 million trainable parameters.

HIGGS

We now test the ability of the polyharmonic cascade to learn from a large-scale dataset with complex nonlinear dependencies and a high level of noise.

For this purpose we used the HIGGS dataset [6, 9], originally introduced in Baldi et al. [6] and available from the UCI Machine Learning Repository [9]. It contains 11 million examples with 28 input features (the first 10.5 million were used for training and the last 500,000 for testing).

Two tests were conducted. In both cases we used a cascade of 20 packages (layers) with architecture 28–200–200–…–200–200–1. All intermediate packages (all except the first and last) thus have 200 inputs and 200 outputs. Initialization (50) was used. The resulting cascade has 21.27 million trainable parameters. The parameter $\alpha$ was set to 1000. The batch size was 14,000.

Test (HIGGS) 1.

In the first test, the training data (in mini-batches) were fed into the cascade in the same form as in the original dataset (no logarithmic transformation of features was performed). Only a min–max normalization was applied so that all input features fell in the range $(-1,1)$.

Training was run for 500 epochs.

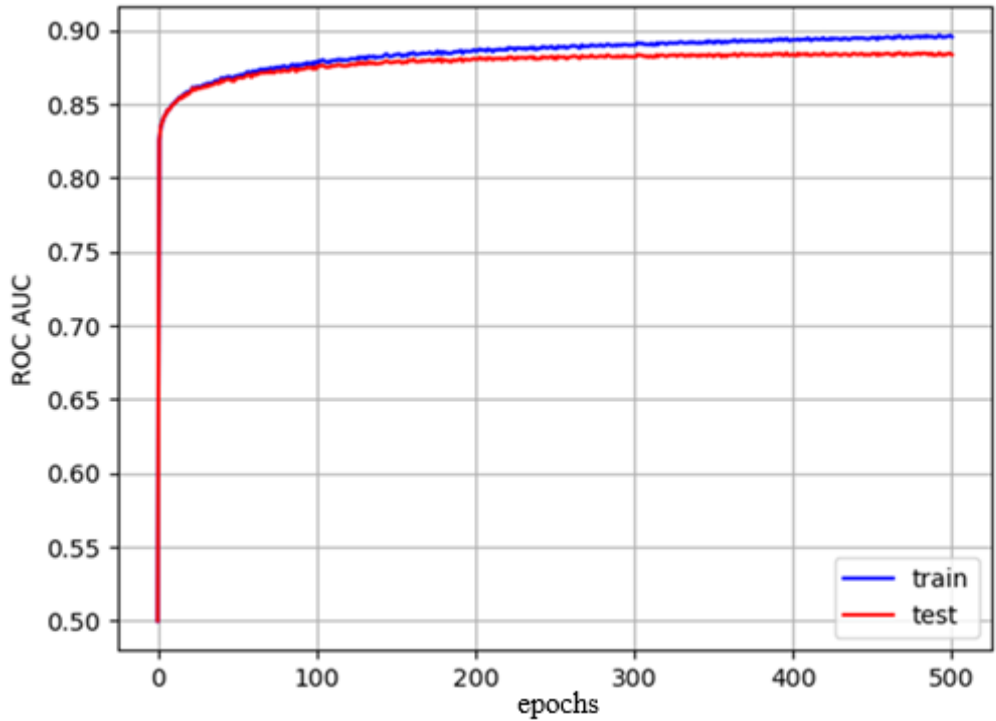

Figure 12. Test 1. Training curve on the HIGGS dataset.

The time per epoch was approximately 14.5 - 15 minutes. As can be seen from the plot, the ROC AUC increases smoothly and in a closely synchronized manner for both the training and test sets throughout the entire training process.

Zooming in on the upper part of the curve:

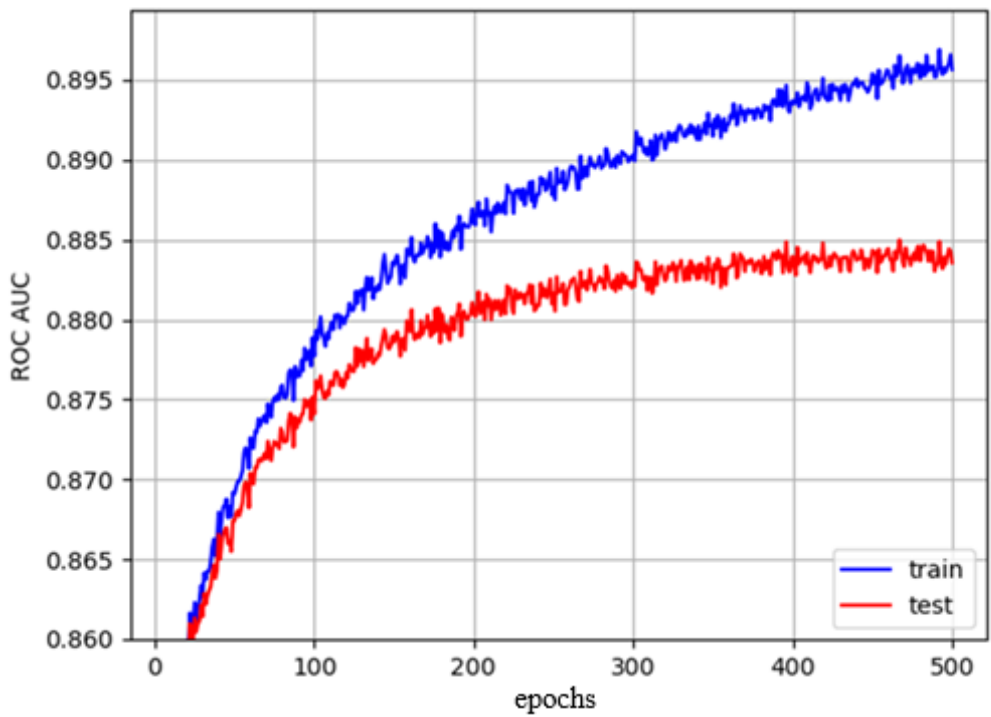

Figure 13. Test 1. Training curve on the HIGGS dataset (zoomed).

We see that an AUC of 0.88 on the test set was reached at about epoch 160. Around epoch 500, the test AUC fluctuates in the range 0.884 - 0.885. The AUC curves for both training and test sets continue to grow (their difference is about 0.012), so it is likely that, if training were continued, the 0.885 mark would be confidently surpassed. No overfitting behavior – where the training AUC continues to increase while the test AUC begins to decrease – is observed over the training horizon shown.

Test (HIGGS) 2.

Test 2 is identical to Test 1 in all respects, except that some of the features were preprocessed. For features with indices 1, 6, 10, 14, 18, 22, 23, 24, 25, 26, 27, and 28 we applied the natural logarithm (NumPy function np.log()). For feature 4 we applied np.log1p(). The goal was to obtain a more uniform distribution of their values. The resulting training curve is shown in Figure 14. At this scale it looks almost indistinguishable from the training curve in Test 1 (Figure 12).

If we zoom in on the upper part of the curve (Figure 15), the plot again looks very similar to Figure 13. However, a closer inspection shows that the change in the distribution of input feature values provides a short-term improvement in learning speed during roughly the first 50 epochs. Later, by epoch 500, the effect becomes the opposite. By the end of training, the test AUC fluctuates around 0.883. Nevertheless, as in Test 1, we see that the AUC continues to increase slowly, and it is plausible that it would eventually reach 0.885 if training were prolonged.

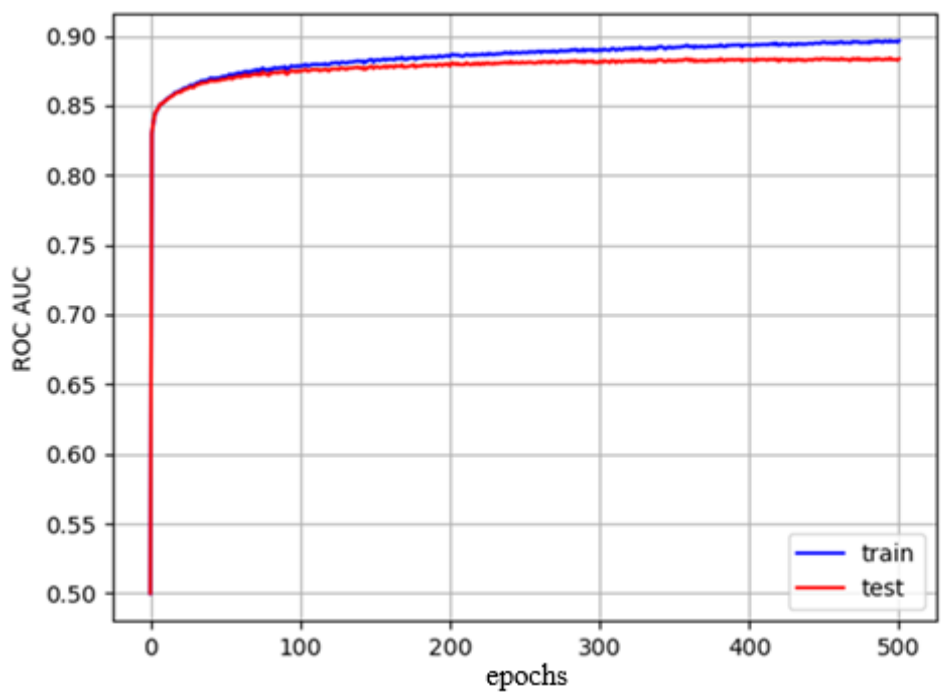

Figure 14. Test 2. Training curve on the HIGGS dataset.

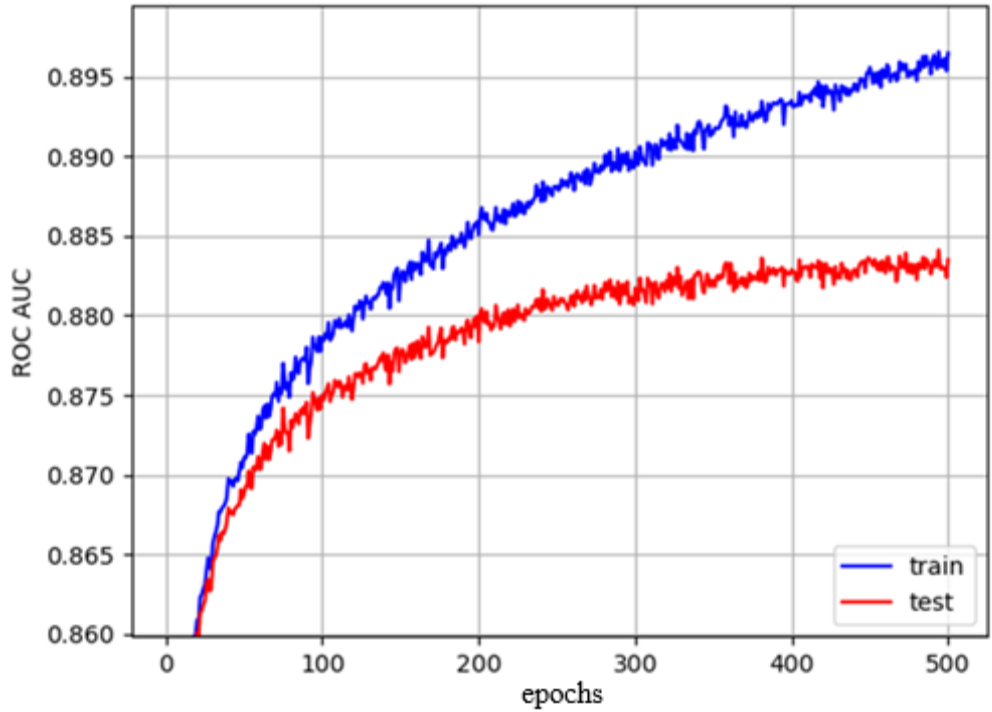

Figure 15. Test 2. Training curve on the HIGGS dataset (zoomed).

In summary, based on the HIGGS experiments, we can conclude that the polyharmonic cascade is capable of handling large datasets with complex nonlinear dependencies and high noise levels, achieving performance comparable to deep neural networks (for example, those reported in [6]).

Epsilon

Next, to evaluate the effectiveness of the polyharmonic cascade on high-dimensional data and large datasets, we used the Epsilon dataset [21] from the PASCAL Large Scale Learning Challenge 2008. This dataset contains 500,000 examples (the first 400,000 were used for training, the remaining 100,000 for testing) with 2000 features. The dataset was loaded via the OpenML API (ID 45575, version 1) using scikit-learn.

Test (Epsilon) 1.

We used a relatively small polyharmonic cascade with architecture 2000–3–20–20–1, i.e., four packages. The total number of trainable parameters in the cascade was 13,004. Training was carried out for 10 epochs. The parameter $\alpha$ was set to 10.

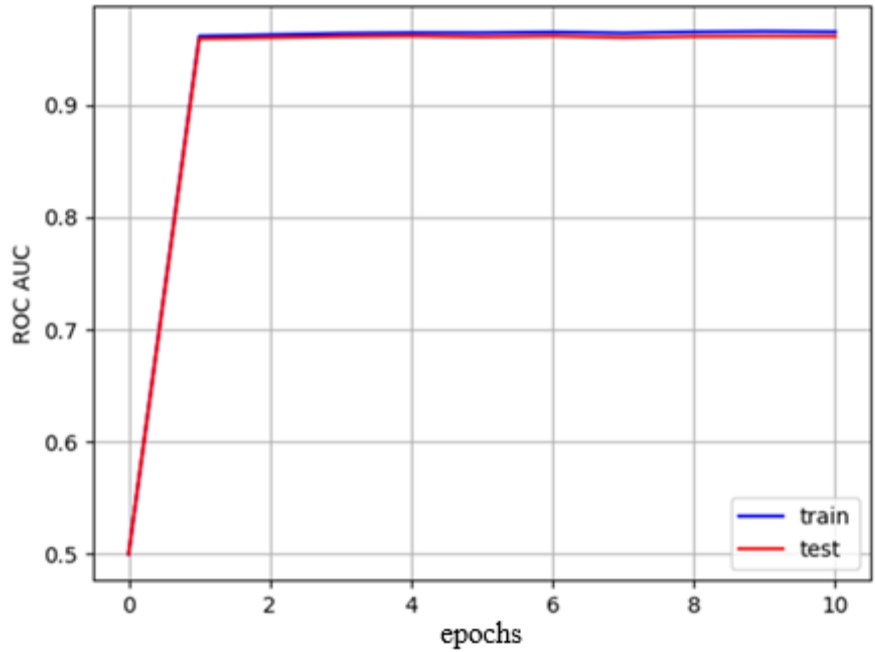

Figure 16. Test 1. Training on the Epsilon dataset.

The training curve on Epsilon resembles the curves obtained on MNIST. The cascade effectively learned already in the first epoch. This test confirmed that the polyharmonic cascade can train on large, high-dimensional datasets and quickly reach high performance.

As training continues, no pronounced overfitting is observed.

Zooming in on the curve:

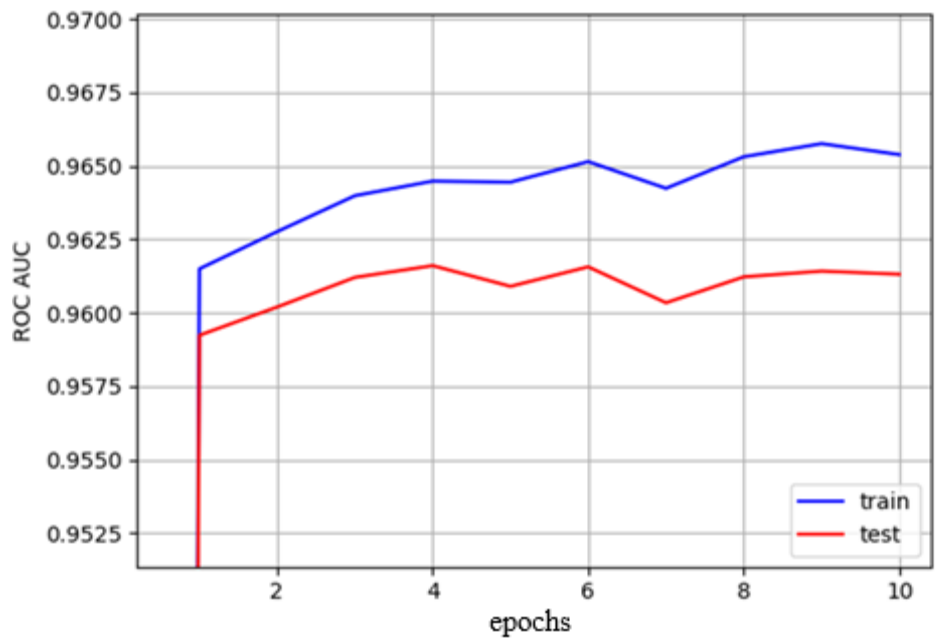

Figure 17. Test 1. Training on the Epsilon dataset (zoomed).

After the first epoch, the ROC AUC reached 0.9592. It then increased to 0.9616 by the fourth epoch and slightly decreased to 0.9613 by the tenth epoch. The time per epoch was just over 5 seconds.

Test (Epsilon) 2.

We now scale up the cascade to architecture 2000–2000–2000–2000–2000–1, using five packages with 2000 inputs and 2000 outputs each (except for the last package). The number of trainable parameters increases to 32 million.

We deliberately choose a large value $\alpha=20000$ to slow down convergence and observe the learning dynamics more carefully. Training was run for 30 epochs.

The time per epoch increased to approximately 37 - 39 seconds. This is about eight times slower than in the previous test, even though the number of trainable parameters in the cascade grew by a factor of roughly 2500.

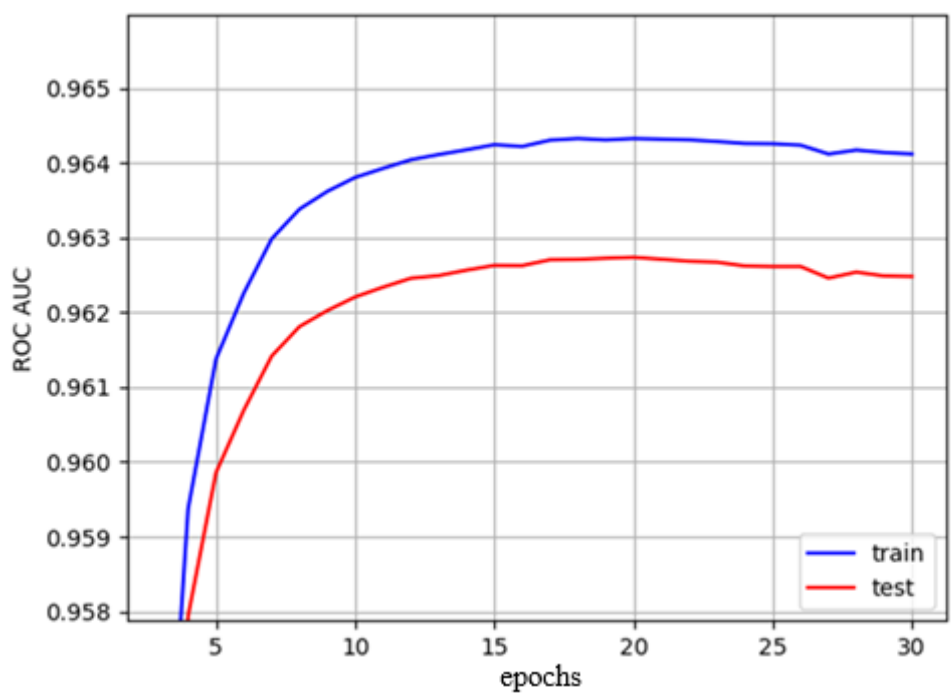

Figure 18. Test 2. Training on the Epsilon dataset.

The ROC AUC reached 0.96 by the fifth epoch, then continued to increase smoothly and plateaued around 0.9627 between epochs 17 and 21. After that, a gradual decline began – an indication that overfitting was starting to manifest – and by epoch 30 the ROC AUC had decreased slightly to 0.9624.

This test confirms that a polyharmonic cascade with a very large number of parameters remains trainable on the Epsilon dataset using exactly the same learning algorithm as for the smaller cascade.

**Conclusion.**

This work completes a cycle of theoretical and applied studies devoted to the polyharmonic cascade – a machine learning architecture whose foundations are derived from first principles (the principle of indifference, symmetry, and Gaussianity). The proposed hyperoctahedron-based initialization ensures both universality and computational efficiency.

The key practical result is the ability to stably train deep cascades (up to 500 layers) without skip connections, normalization, or other heuristics that are widely used in neural networks to combat vanishing gradients. This indicates a fundamental robustness of the proposed architecture, rooted in its probabilistic nature.

The results obtained on the HIGGS and Epsilon benchmarks confirm the practical viability of the polyharmonic cascade concept and the soundness of the proposed paradigm, which was the primary goal of this research.

Although at this early stage (the very first experiments with the new architecture) the objective was not to compete with state-of-the-art (SOTA) methods, we can still provisionally place the results among modern machine learning approaches. It is important to emphasize that all experiments were conducted without sophisticated feature engineering, without extensive hyperparameter tuning, and without specialized techniques such as ensembling or multi-stage training.

On the Epsilon dataset (500,000 examples, 2000 features), the polyharmonic cascade achieved ROC AUC ≈ 0.9627. This result should be viewed in the context of other models evaluated under comparable conditions. For example, in a one-pass learning setting, the OPAUC algorithm achieved AUC 0.9550±0.0007 [10], Large-scale Tree Boosting reached AUC ≈ 0.95 [20], and classical methods such as Naive Bayes, even after feature selection, did not exceed AUC 0.79 [19]. Moreover, the performance of the polyharmonic cascade lies very close to the SOTA result reported for CatBoost in 2020 (AUC = 0.964) [12]. Considering that CatBoost is a mature framework with multi-level optimizations, whereas the polyharmonic cascade is presented here in its first, basic implementation, this proximity highlights the high effectiveness of the proposed architecture in settings with high dimensionality and noise.

On the HIGGS dataset (full dataset, 11 million examples, 28 features), the cascade achieved ROC AUC ≈ 0.885. This substantially outperforms baseline models: logistic regression (~0.64), decision tree (~0.63), random forest (~0.67), and gradient boosted trees (0.705) [2]. It also exceeds the performance of early neural models such as the perceptron (AUC 0.668) and a DNN with three hidden layers (AUC 0.842) [16]. The result is comparable to those reported in the original paper [6], as well as to the performance of a CNN applied to visualized event data (AUC

0.904) [1] – which is particularly noteworthy given that in our experiments we worked with tabular data, without transforming them into images.

We can therefore conclude that the polyharmonic cascade does not merely demonstrate "conceptual feasibility." It is competitive with many standard and even advanced methods on challenging, widely used benchmarks. These results provide strong motivation for further investigation of this paradigm, including its potential in combination with feature engineering, architecture optimization, improved initialization algorithms, and applications to other classes of problems.

Thus, the paper not only resolves the question of initialization, but also demonstrates the coherence of the entire framework: from first-principles derivation of the kernel to a scalable implementation of a deep architecture.

**Acknowledgments.**

The author thanks the anonymous reviewers for valuable feedback on earlier versions of this work.